\documentclass{article}

\usepackage[nonatbib,preprint]{neurips_2023}

\usepackage[utf8]{inputenc} 
\usepackage[T1]{fontenc}    
\usepackage{hyperref}       
\usepackage{url}            
\usepackage{booktabs}       
\usepackage{amsfonts}       
\usepackage{nicefrac}       
\usepackage{microtype}      
\usepackage{xcolor}         

\usepackage[style=numeric-comp,sorting=none,maxcitenames=2]{biblatex}
\usepackage[acronym,shortcuts]{glossaries}
\usepackage{xspace}
\usepackage{graphicx}
\usepackage{subcaption}
\usepackage[capitalise]{cleveref}
\usepackage{amsmath}
\usepackage{bm}
\usepackage{pifont}
\usepackage{scalefnt}
\usepackage{adjustbox}
\usepackage{pgf}
\usepackage{import}
\usepackage{enumitem}

\usepackage{wrapfig}

\definecolor{figred}{HTML}{B85450}
\definecolor{figyellow}{HTML}{D79B00}
\definecolor{figgreen}{HTML}{82B366}
\definecolor{figblue}{HTML}{6C8EBF}
\definecolor{figpink}{HTML}{FF33FF}

\newcommand\smaller[2][0.8]{{\scalefont{#1}#2}}

\newcommand{\xmark}{\ding{55}}

\glsdisablehyper

\makeatletter
\DeclareRobustCommand\onedot{\futurelet\@let@token\@onedot}
\def\@onedot{\ifx\@let@token.\else.\null\fi\xspace}
 
\def\ie{\emph{i.e}\onedot}

\makeatother

\newacronym{wsi}{WSI}{whole slide image}
\newacronym{he}{H\&E}{hematoxylin and eosin}
\newacronym{gnn}{GNN}{graph neural network}
\newacronym{mil}{MIL}{multiple instance learning}
\newacronym{cnn}{CNN}{convolutional neural network}
\newacronym{auroc}{AUROC}{area under the receiver operating characteristic}

\crefname{align}{}{}

\def\zhat{\hat{z}}

\def\vzero{{\bm{0}}}

\def\vb{{\bm{b}}}

\def\vg{{\bm{g}}}
\def\vh{{\bm{h}}}

\def\vp{{\bm{p}}}

\def\vr{{\bm{r}}}

\def\vu{{\bm{u}}}
\def\vv{{\bm{v}}}
\def\vw{{\bm{w}}}
\def\vx{{\bm{x}}}

\def\vz{{\bm{z}}}

\def\vzhat{{\bm{\hat{z}}}}

\def\mA{{\bm{A}}}

\def\mD{{\bm{D}}}

\def\mW{{\bm{W}}}
\def\mX{{\bm{X}}}

\DeclareMathAlphabet{\mathsfit}{\encodingdefault}{\sfdefault}{m}{sl}
\SetMathAlphabet{\mathsfit}{bold}{\encodingdefault}{\sfdefault}{bx}{n}

\newcommand{\R}{\mathbb{R}}

\bibliography{bibliography}

\title{Deep Multiple Instance Learning with Distance-Aware Self-Attention}

\author{%
  Georg Wölflein\\
  School of Computer Science\\
  University of St Andrews\\
  \texttt{georg@woelflein.de} \\
  \And
  Lucie Charlotte Magister \\
  Department of Computer Science and Technology\\
  University of Cambridge \\
  \texttt{lcm67@cam.ac.uk} \\
  \And
  Pietro Li\`{o} \\
  Department of Computer Science and Technology\\
  University of Cambridge \\
  \texttt{pl219@cam.ac.uk} \\
  \And
  David J.\ Harrison \\
  School of Medicine\\
  University of St Andrews\\
  \texttt{david.harrison@st-andrews.ac.uk} \\
  \And
  Ognjen Arandjelovi\'c \\
  School of Computer Science\\
  University of St Andrews\\
  \texttt{oa7@st-andrews.ac.uk}\\
}

\AtBeginDocument{
  \addtolength{\abovedisplayskip}{-3pt}
  \addtolength{\abovedisplayshortskip}{-3pt}
  \addtolength{\belowdisplayskip}{-3pt}
  \addtolength{\belowdisplayshortskip}{-3pt}
}

\begin{document}

\maketitle

\renewcommand{\baselinestretch}{0.98}

\begin{abstract}
  Traditional supervised learning tasks require a label for every instance in the training set, but in many real-world applications, labels are only available for collections (bags) of instances.
  This problem setting, known as multiple instance learning (MIL), is particularly relevant in the medical domain, where high-resolution images are split into smaller patches, but labels apply to the image as a whole.
  Recent MIL models are able to capture correspondences between patches by employing self-attention, allowing them to weigh each patch differently based on all other patches in the bag.
  However, these approaches still do not consider the relative spatial relationships between patches within the larger image, which is especially important in computational pathology.
  To this end, we introduce a novel MIL model with distance-aware self-attention (DAS-MIL), which explicitly takes into account relative spatial information when modelling the interactions between patches.
  Unlike existing relative position representations for self-attention which are discrete, our approach introduces continuous distance-dependent terms into the computation of the attention weights, and is the first to apply relative position representations in the context of MIL.
  We evaluate our model on a custom MNIST-based MIL dataset that requires the consideration of relative spatial information, as well as on CAMELYON16, a publicly available cancer metastasis detection dataset, where we achieve a test AUROC score of 0.91.
  On both datasets, our model outperforms existing MIL approaches that employ absolute positional encodings, as well as existing relative position representation schemes applied to MIL.
  Our code is available at \url{https://anonymous.4open.science/r/das-mil}.
\end{abstract}

\section{Introduction}

\Gls{mil}~\cite{dietterich1997solving,maron1997framework} is a supervised learning framework for data that is supplied as bags of instances, where only the bag labels are known. It has been used in various domains such as image classification~\cite{wu2015deep}, object detection~\cite{wan2019c}, drug discovery~\cite{fu2012implementation}, and text categorization~\cite{pappas2014explaining}. 
Recently, \gls{mil} has attracted a lot of attention in the medical imaging community, where it has become the predominant paradigm for classifying \glspl{wsi}~\cite{campanella2019clinical,niehues2023generalisable,saldanha2023self,elnahhas2023regression}.
These are high-resolution, digitised microscopic images of tissue sections which, due to their large size, need to be divided into smaller patches to be processed by deep learning models.
Labels are only available at the bag level, \ie for the entire \gls{wsi}, and not for individual instances (patches).

\begin{figure}[th]
  \centerline{\resizebox{0.75\textwidth}{!}{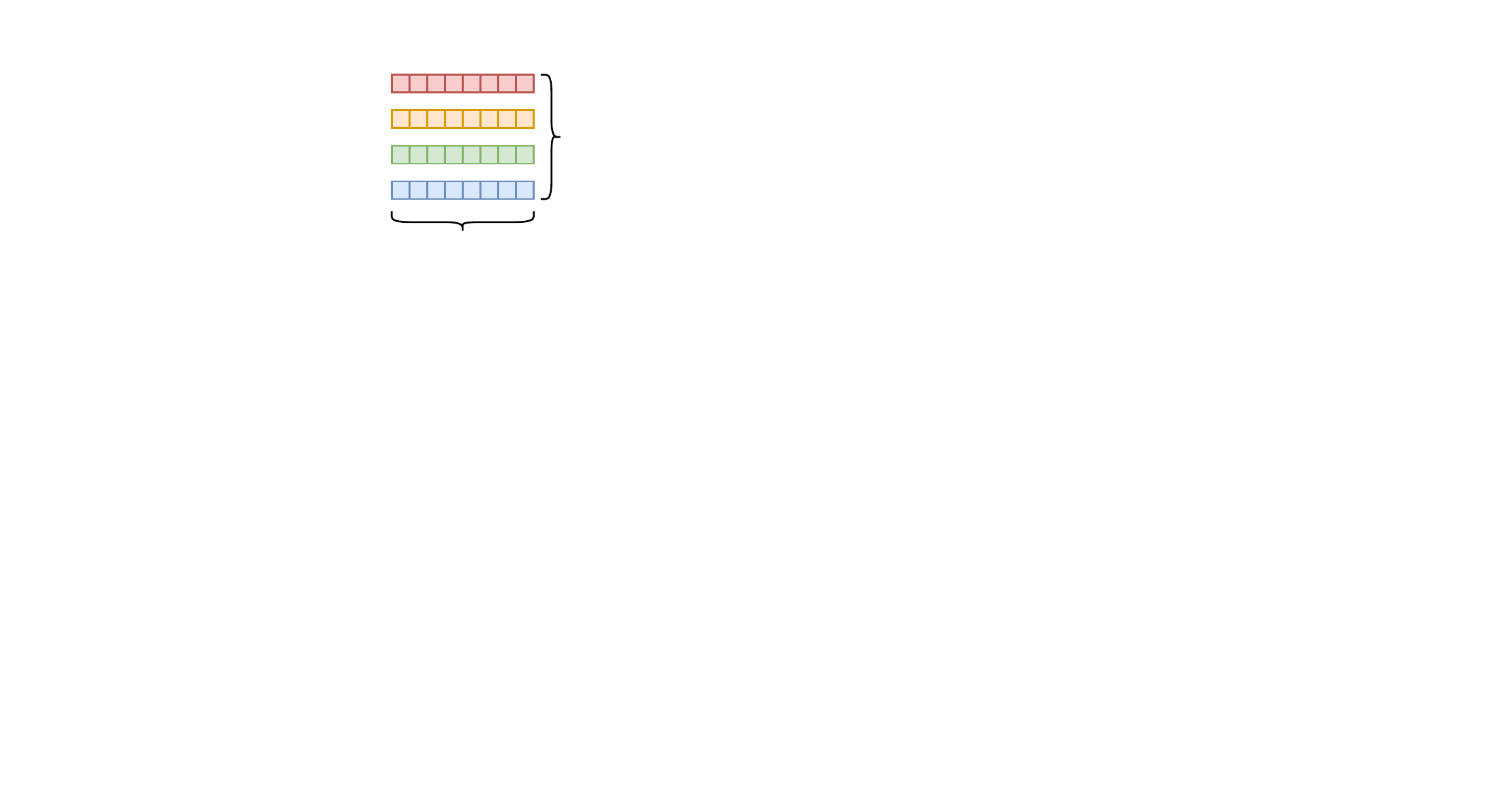}}
  \caption{We propose a \acs{mil} pipeline (top) with a novel self-attention mechanism (bottom), incorporating relative distance information (\textcolor{figpink}{magenta}) in the computation of the attention weights.}
  \label{fig:overview}
  \vspace{-12pt}
\end{figure}

A standard \gls{mil} assumption in binary classification is that at least one instance per positive bag has the target property, while all instances in negative bags lack it.
For example, in classifying \glspl{wsi} as cancerous or non-cancerous, the assumption is that a cancerous \gls{wsi} has at least one patch containing tumour tissue, while a non-cancerous \gls{wsi} does not contain any tumour tissue.
A na\"ive approach is to apply a standard classifier to each instance independently and then aggregate the predictions to obtain the bag-level prediction.
However, previous work has shown that this approach runs into the problem of insufficiently training the instance-level classifier due to incorrect instance-level labels~\cite{wang2018revisiting,ilse2018attention}.
Instead, we focus on the generally preferred embedding-level approach, where the instances are first embedded into a common space and then aggregated to obtain the bag-level prediction~\cite{wang2018revisiting,ilse2018attention}.

Embedding-level \gls{mil} models~\cite{wang2018revisiting,ilse2018attention} still operate on the classical \gls{mil} assumption that the instances within a bag are independent and that the bag-level prediction can be obtained by aggregating the instance-level predictions.
However, this assumption does not always hold in real-world applications such as medical imaging, where instances may exhibit complex dependencies.

Self-attention mechanisms \cite{vaswani2017attention} have shown great success in various natural language processing and computer vision tasks because they are able to weigh parts of an input sequence differently based on the entire sequence.
In an attempt to weaken the independence assumption in \gls{mil}, recent works have proposed to use self-attention in the aggregation step~\cite{rymarczyk2021kernel,chen2021multimodal,myronenko2021accounting,li2020dual,li2020deep,xiong2023diagnose,shao2021transmil,wagner2023fully,huang2021integration,chen2022scaling,qian2022transformer,ding2022deep} because this allows the model to weigh each instance differently based on all other instances in the bag.

In this paper, we study a special case of \gls{mil} where instances are equipped with a distance metric.
This problem setting arises naturally in the medical imaging domain, where high-resolution histological images are split into smaller patches, and distances between patches can be defined as the Euclidean distance between their centroid coordinates.
However, current \gls{mil} models, even those employing self-attention in the embedding space~\cite{rymarczyk2021kernel,chen2021multimodal,myronenko2021accounting,li2020dual,li2020deep,xiong2023diagnose,shao2021transmil,wagner2023fully,huang2021integration,chen2022scaling,qian2022transformer,ding2022deep}, do not consider the relative distances between instances, which may be important for the classification task. 
For example, in the computational pathology domain, the spatial relationships between various tissue structures have been shown to be of prognostic value, reflecting the importance of direct cell-cell interactions in influencing biological behaviour~\cite{nearchou2019automated,todd2019identifying,defilippis2022use}.
Furthermore, the interaction between tumour cells and immune cells, which is inherently spatial in nature, is a factor that may be pivotal in predicting the response to immunotherapy~\cite{hackl2016computational,shelton2021engineering}.

To address this problem, we introduce a novel \gls{mil} model with distance-aware self-attention (DAS-MIL), which explicitly takes into account the pairwise distances between instances in a bag when computing the attention weights (see \cref{fig:overview}). 
To the best of our knowledge, our approach is the first to use relative position representations in \gls{mil}, and in the context of relative position representations, the first to use a continuous distance-dependent bias term in the self-attention mechanism.
We evaluate our model on a custom dataset requiring distance awareness for prediction, as well as \smaller{CAMELYON16}~\cite{bejnordi2017diagnostic}, a publicly available dataset for cancer metastasis detection. 

The remainder of this paper is organized as follows. 
In \cref{sec:background}, we review related work, before we introduce our DAS-MIL model in \cref{sec:method}.
In \cref{sec:results}, we then report experimental results, before we conclude the paper in \cref{sec:conclusion} and discuss future directions of research.

\section{Background and Related Work}
\label{sec:background}

\subsection{Multiple instance learning}
\label{sec:mil}
In the binary \gls{mil} setting, the goal is to predict the label $Y \in \{0,1\}$ of a bag $\mX$ of instances $\{\vx_1, \vx_2, \dots, \vx_n\}$ with each instance $\vx_i \in \R^n$.
Note that the number of instances in a bag, $n$, may vary.
The challenge is that only the bag-level label $Y$ is known and not the corresponding instance-level labels $\{y_{1}, y_{2}, \dots, y_{n}\}$, each $y_i \in \{0,1\}^{n}$.
However, it can be assumed that there exists at least one positive instance per positive bag, while all instances in negative bags are negative, \ie,
\begin{align}
  \label{eq:mil_assumption}
  Y = \begin{cases}
    1 & \text{iff} \ \sum\limits_{i=1}^{n_i}{y_i} > 0 \\
    0 & \text{otherwise}
  \end{cases}.
\end{align}
Our training dataset consists of $N$ bags $\mX^{(i)}$ with corresponding labels $Y^{(i)}$ for $i=1,2,\dots,N$, where each bag $\mX^{(i)}$ contains $n_i$ instances $\{\vx^{(i)}_1, \vx^{(i)}_2, \dots, \vx^{(i)}_{n_i}\}$.
However, to ease notation, we omit the bag index $i$, as it is clear from the context that we are referring to some bag $\mX$ with associated label $Y$. 
It is worth noting, that originally instances within each bag were assumed to be independent and identically distributed (i.i.d.)~\cite{dietterich1997solving,maron1997framework}, but recent works have recognised this assumption to often be violated in real-world applications~\cite{shao2021transmil,rymarczyk2021kernel}.

A \gls{mil} learns a scoring function $S : \R^{n \times d_x} \to \R$ that predicts the bag label $Y$ from the bag $\mX$:
\begin{align}
  \label{eq:mil_score}
  \hat{Y} = S(\mX)
\end{align}
We can decompose the scoring function into three parts.
First, a transformation $\vh : \R^{d_x} \to \R^{d_z}$ computes an embedding vector $\vz_i$ of dimensionality $d_z$ for each instance $\vx_i$ in our bag $\mX$.
Then, an aggregation function $\vg : \R^{n \times d_z} \to \R^{d_z}$ combines all the embeddings to a single embedding vector which is finally passed through a classifier $f : \R^{d_z} \to \R$ to obtain the bag-level prediction:
\begin{align}
  \label{eq:mil_score_embedded:embedding}
  \vz_{i} &= \vh(\vx_i)\\
  \label{eq:mil_score_embedded:aggregation}
  \vz' &= \vg\left(
    \left\{\vz_1, \vz_2, \dots, \vz_n\right\}
  \right)\\
  \label{eq:mil_score_embedded:score}
  S(\mX) &= f\left(\vz'\right).
\end{align}
As a basic example, one could parameterize $f$ and $\vh$ using neural networks and employ the mean function as the aggregation function $\vg$.

\vspace{-8pt}\paragraph{Attention-based MIL}
A popular approach to \gls{mil} known as att-MIL employs a simple attention mechanism in the aggregation function~\cite{ilse2018attention}.
Instead of using the mean function, att-MIL independently computes an individual attention weight $\alpha_i$ for each instance $\vx_i$ in the bag $\mX$, and then aggregates the embeddings as a weighted average.
However, this approach still assumes that instances are independent because the attention weights are computed independently for each instance.

\subsection{Self-attention}
\label{sec:self_attention}
The self-attention mechanism was originally introduced by \textcite{vaswani2017attention} as a tool for modelling long-range dependencies in sequential data, though it can equally be applied to permutation-invariant data, 
such as instances in a bag.
An attention head operates on a set of input vectors $\{\vx_1, \vx_2, \dots, \vx_n\}$ where each $\vx_i\in\R^{d_x}$ and computes a set of the same number of output vectors $\{\vz_1, \vz_2, \dots, \vz_n\}$, each $\vz_i\in\R^{d_z}$. 
However, the dimensionality of the input vectors $d_x$ and output vectors $d_z$ may differ.

A particular output vector $\vz_i$ is computed as a weighted sum of linearly transformed input vectors $\vx_j$,
\begin{align}
  \label{eq:self_att}
  \vz_i = \sum_{j=1}^n{\alpha_{ij}(\vx_j \mW^V)},
\end{align}
where the linear transformation is parameterised by a weight matrix $\mW^V\in\R^{d_x \times d_z}$.
The coefficients in the weighted sum $\alpha_{ij}$ are computed as a softmax over the compatibilities $e_{ij}$.
The term $e_{ij}$ expresses the compatibility between input vectors $\vx_i$ and $\vx_j$, and is computed as a dot product between the two vectors after each is transformed by a separate linear function:
\begin{align}
  \label{eq:self_att_compatibility}
  e_{ij} = \frac{(\vx_i \mW^Q)(\vx_j \mW^K)^\top}{\sqrt{d_z}}.
\end{align}
These linear transformations are parameterized by two more weight matrices $\mW^Q,\mW^K\in\R^{d_x \times d_z}$.
Intuitively, $\vx_j$ undergoes a linear transformation via $\mW^K$ to emit a key vector, which is compared to the query vector obtained by transforming $\vx_i$ via $\mW^Q$.

\vspace{-8pt}\paragraph{Positional and spatial information}
On its own, the self-attention mechanism is permutation invariant.
For sequential data (one-dimensional in the sense that there is only a temporal dimension), this issue was already addressed in the original transformer model~\cite{vaswani2017attention} by adding a positional encoding $\vp_i\in\R^{d_x}$ to each input vector $\vx_i$ before applying the self-attention mechanism.
On the other hand, \textcite{shaw2018self} propose using relative position representations by adding learned representations to the key and value vectors of \cref{eq:self_att,eq:self_att_compatibility} that depend on the relative distances between input tokens.
\Textcite{liutkus2021relative} take this idea further to make it linear in complexity.

The 2D case, which is the focus of our paper, is more challenging because the input vectors $\vx_i$ are not ordered in a sequence, but rather associated with 2D coordinates.
A simple approach is to perform the original positional encoding~\cite{vaswani2017attention} separately on the $x$ and $y$ coordinates of each input element, and to then concatenate the resulting vectors~\cite{carion2020end,dosovitskiy2021image,parmar2018image,huang2021integration}.
However, the dot product of two such positionally encoded vectors is no longer proportional to the $\ell_2$ distance between the two locations, meaning that this information is lost.
The solution proposed by \cite{yang2021learnable} is a learned relative positional encoding using Fourier features, which retains the $\ell_2$ distance information in the dot product.

However, instead of using a form of absolute 2D positional encoding, our approach directly incorporates relative distance information in the self-attention mechanism by adding a learned distance-dependent bias term to the attention weights in the compatibility function~\cref{eq:self_att_compatibility}.
In the literature, approaches involving a bias term in the attention weights have only been studied in the context of discrete distances~\cite{ying2021transformers,wu2021rethinking}, while we consider continuous distances.
In graph representation problems, \textcite{ying2021transformers} add a bias to the attention weights in the transformer using a trainable scalar.
Perhaps the most similar approach to ours is that of \textcite{wu2021rethinking}, who propose variations of relative positional representations~\cite{shaw2018self} for patch-based self-attention in image classification.
Nonetheless, their approach is limited to discrete distance values because they split the input image into a regular grid of patches, while we consider patches where there are no such restrictions.

\subsection{Self-attention in multiple instance learning}
As mentioned earlier, many practical \gls{mil} problems require the consideration of more than one instance in a bag for a decision to be made which is in contrast to the original assumption in \cref{eq:mil_assumption}.
Consequently, a new class of \gls{mil} models is emerging where the i.i.d.\ assumption of the patches is relaxed by an aggregation function that has the ability to consider dependencies between instances in the bag. The self-attention mechanism is particularly well-suited for this task, however, other models such as \glspl{gnn}~\cite{tu2019multiple} have also been considered in this context.

A number of recent \gls{mil} models employ self-attention in the aggregation function~\cite{rymarczyk2021kernel,chen2021multimodal,myronenko2021accounting,li2020dual,li2020deep,xiong2023diagnose,shao2021transmil,wagner2023fully,huang2021integration,chen2022scaling,qian2022transformer,ding2022deep} to capture dependencies between input tokens in a sequence, which in the case of \gls{mil} are instances in a bag.
Interestingly, almost all of them are designed to operate on image patches, which is the focus of our work.
These approaches consistently outperform classical \gls{mil} models operating under the patch independence assumption.
Some approaches regard a bag simply as a set of image patches, ignoring their spatial arrangement altogether~\cite{rymarczyk2021kernel,chen2021multimodal,myronenko2021accounting,li2020dual,li2020deep}, however, this information may be crucial, especially in the medical domain.
Approaches that do incorporate spatial information typically employ absolute positional encodings~\cite{shao2021transmil,wagner2023fully,huang2021integration,chen2022scaling,qian2022transformer,ding2022deep}, though recently there has been work on using positional hierarchies~\cite{xiong2023diagnose}.
However, to the best of our knowledge, there has been no work on using relative positional representations in \gls{mil}, which our focus herein.

\section{DAS-MIL}
\label{sec:method}

Our model follows the three-step embedding-based \gls{mil} approach described in \cref{sec:mil}. First, 
We parameterise the feature extractor $\vh$ in \cref{eq:mil_score_embedded:embedding} using a \gls{cnn} whose flattened outputs pass through a linear layer with $d_z$ output units. Each image patch is processed independently by the \gls{cnn} to produce a feature vector $\vz_i\in\R^{d_z}$ for each patch $i$.
Secondly, we apply a novel type of self-attention which we term \emph{distance-aware self-attention}, described in the next section, to the $\vz_i$ to obtain a new set of feature vectors $\vz'_i\in\R^{d_z}$,
\begin{align}
    \label{eq:method:das_att}
    \{\vz'_1, \vz'_2, \dots, \vz'_n\} = \mathrm{DASAtt}(\{\vz_1, \vz_2, \dots, \vz_n\}, \mD)
  \end{align}
  based on the distance matrix $\mD\in\R^{n\times n}$ whose $(i,j)$-th entry is the distance between the $i$-th and $j$-th patch $\delta_{ij}$.
  Then, we aggregate the new feature vectors $\vz'_i$ to a single vector $\vzhat \in \R^{d_z}$ by computing the maximum value along each dimension such that the $j$-th element of $\vzhat$ is given by
  \begin{align}
    \label{eq:method:z_prime}
    \zhat_j = \max_{1 \leq i \leq n}{z'_{i,j}}
  \end{align}
  where $z'_{i,j}$ is the $j$-th component of $\vz'_i$.
  Together, \cref{eq:method:das_att,eq:method:z_prime} form the aggregation function $\vg$ in \cref{eq:mil_score_embedded:embedding}.
Finally, we use the aggregated feature vector $\vzhat$ to compute the \gls{mil} score using a linear layer with a single output unit. Specifically, we parameterise the scoring function $f$ in \cref{eq:mil_score_embedded:embedding} as
  \begin{align}
    \label{eq:method:score}
    S(\mX) = f(\vzhat) = \sigma\left(
      \vzhat \vw^S + b^S
    \right)
  \end{align}
  with weight and bias parameters $\vw^S\in\R^{d_z}$ and $b^S\in\R$, respectively, and sigmoid activation $\sigma$.

We train the model end-to-end using a weighted version of the binary cross-entropy loss that accounts for class imbalances in the training set.
The model is presented one bag at a time (\ie in a batch of size one) with the loss for a particular bag $\mX$ and associated label $Y$ computed as:
\begin{align}
  \label{eq:method:loss}
  \mathcal{L} = -\omega Y\log S(\mX) - (1-Y)\log(1-S(\mX))
\end{align}
where the weight $\omega\in\R$ is the ratio of negative to positive bags in the training set.

\subsection{Distance-aware self-attention}
\label{sec:distance_aware_self_attention}

The central component of our model is the distance-aware self-attention mechanism.
We retain the notation from \cref{sec:self_attention} in that the self-attention mechanism acts on a set of input vectors $\vx_1,\vx_2,\dots,\vx_n$, transforming them into a set of output vectors $\vz_1,\vz_2,\dots,\vz_n$, which is consistent with the literature~\cite{vaswani2017attention,shaw2018self,wu2021rethinking}.
This is not to be confused with the notation in \cref{sec:method}, where the input vectors to the self-attention mechanism are embedding vectors, already denoted as $\vz_i$ by convention in the \gls{mil} literature~\cite{ilse2018attention}; thus the output vectors were denoted as $\vz'_i$ above. 

We first describe a general framework for incorporating relative distance information in self-attention, and then introduce our method as an instantiation of this framework in the next section. We then compare our method to existing work~\cite{shaw2018self,dai2019transformer,huang2020improve,ramachandran2019stand,wang2020axial,wu2021rethinking} in \cref{sec:method:prior}.

The general framework introduces three terms into the self-attention mechanism, $\vb^K_{ij},\vb^Q_{ij}\in\R^{d_x}$ and $\vb^V_{ij}\in\R^{d_z}$.
Intuitively, these are learned representations of the relative distances between the $i$-th and $j$-th input elements.
We add $\vb^V_{ij}\in\R^{d_x}$ in the computation of the output vector $\vz_i$ from \cref{eq:self_att} as
\begin{align}
  \label{eq:das:output}
  \vz_i = \sum_{j=1}^n{\alpha_{ij}\left(\vx_j \mW^V \color{blue} + \vb^V_{ij}\normalcolor\right)}.
\end{align}
In the compatibility function in \cref{eq:self_att_compatibility}, we add $\vb_{ij}^K$ to the key vector $\vx_j \mW^K$ in order to infuse distance information into the computation of the attention weights (and analogously $\vb_{ij}^Q$):
\begin{align}
  \label{eq:das:compatibility}
  e_{ij} = \frac{(\vx_i \mW^Q \color{blue} + \vb^Q_{ij}\normalcolor)(\vx_j \mW^K \color{blue} + \vb^K_{ij}\normalcolor)^\top \color{blue} - \vb^Q_{ij} {\vb^K_{ij}}^\top\normalcolor}{\sqrt{d_z}}.
\end{align}

\vspace{-8pt}\paragraph{Efficient implementation}
There are two reasons for subtracting the term $\vb^Q_{ij} {\vb^K_{ij}}^\top$ in \cref{eq:das:compatibility}.
It improves the performance of our model (as we show in an ablation study) and saves some computation in the implementation of the compatibility function, as we explain next.
When we rewrite \cref{eq:das:compatibility} as
\begin{align}
  \label{eq:das:compatibility:impl}
  e_{ij} &= \frac{(\vx_i \mW^Q) (\vx_j \mW^K)^\top + (\vx_i \mW^Q) \color{blue}{\vb^K_{ij}}^\top\normalcolor + (\vx_j \mW^K)^\top \color{blue} \vb^Q_{ij}\normalcolor}{\sqrt{d_z}},
\end{align}
the term $\vb^Q_{ij} {\vb^K_{ij}}^\top$ cancels out.
However, more importantly, we can precompute $(\vx_i \mW^Q) (\vx_j \mW^K)^\top$ for all pairs of input elements $i,j$ as in the original self-attention mechanism~\cite{vaswani2017attention}, and store it in a matrix $\mA \in\R^{n\times n}$.
First, we calculate the matrix $\mX \mW^Q \in \R^{n\times d_z}$, where $\mX \in \R^{n\times d_x}$ is the matrix whose rows are the input vectors $\vx_i$.
Similarly, we compute $\mX \mW^K \in \R^{n\times d_z}$.
This gives rise to $\mA=(\mX \mW^Q)(\mX \mW^K)^\top$ whose $(i,j)$-th element is the first term in \cref{eq:das:compatibility:impl}.
Furthermore, we can reuse the matrices $\mX \mW^Q$ and $\mX \mW^K$ in the computation of the other two terms.

\subsubsection{Our method}

Let $\delta_{ij}$ be the Euclidean distance between instances $i$ and $j$.
We define the relative positional encoding for the key vectors $\vb^K_{ij}$ as an interpolation between two learned vectors $\vu^K,\vv^K \in \R^{d_x}$:
\begin{align}
  \label{eq:bijK}
  \vb_{ij}^K = \phi(\delta_{ij})\vu^K + (1-\phi(\delta_{ij}))\vv^K,
\end{align}
where the interpolation coefficient $\phi(\delta)$ is obtained via a learned function $\phi : \R \to \R$ that maps the relative distance $\delta$ to a scalar.
We parameterise $\phi$ as a sigmoid function that is scaled and shifted by learnable parameters $\beta,\theta \in \R$.
Similarly, we define the encodings for $\vb^Q_{ij}$ and $\vb^V_{ij}$ as
\begin{align}
  \label{eq:bijQV}
  \vb_{ij}^Q = \phi(\delta_{ij})\vu^Q + (1-\phi(\delta_{ij}))\vv^Q &&\text{and}&&
  \vb_{ij}^V = \phi(\delta_{ij})\vu^V + (1-\phi(\delta_{ij}))\vv^V,
\end{align}
using different trainable vectors $\vu^Q,\vv^Q \in \R^{d_x}$ and $\vu^V,\vv^V \in \R^{d_z}$.
However, we find it beneficial to share the parameters $\beta$ and $\theta$ of the $\phi$ function between all three encoding vectors $\vb^K_{ij},\vb^Q_{ij},\vb^V_{ij}$.

More formally, the procedure outlined in this section gives rise to the distance-aware self-attention operator as it is used in \cref{eq:method:das_att}, which computes a set of output vectors $\vz_i$ based on the input vectors $\vx_i$ and their relative distances $\delta_{ij}$ packed into a distance matrix $\mD\in\R^{n\times n}$:
\begin{align}
  \mathrm{DASAtt}\left(\left\{\vx_1,\vx_2,\ldots,\vx_n\right\}, \mD\right) = \left\{\vz_1,\vz_2,\ldots,\vz_n\right\}.
\end{align}

\subsubsection{Prior work}
\label{sec:method:prior}
Adding distance dependent terms to the key vectors in \cref{eq:das:compatibility} and the output vectors in \cref{eq:das:output} was first proposed by \textcite{shaw2018self} in the context of machine translation.
The authors learn a relative position representation vector $\vr_k^K \in \R^{d_z}$ for every relative distance $k\in\mathbb{Z}$ between input token positions (up to some maximum distance $k_\mathrm{max}$ such that $\lvert k \rvert \leq k_\mathrm{max}$) and set $\vb^K_{ij} = \vr_{i-j}^K$.
The same procedure is followed for $\vb^V_{ij}$, with the difference that the query vector is not altered, \ie $\vb^Q_{ij}=\vzero$.
However, note that the original work~\cite{shaw2018self}, as well as some of the other methods that followed~\cite{dai2019transformer,huang2020improve} (a comprehensive overview is provided by \textcite{wu2021rethinking}), study the 1D case of language modelling where relative distances are discrete values representing token offsets.

There is some prior work on extending relative position representations to 2D images, where the image is split into a regular grid of tiles constituting the input elements. 
One approach is to compute separate relative position representations in the horizontal and vertical directions, and then either to concatenate them~\cite{ramachandran2019stand} or use them sequentially in consecutive layers~\cite{wang2020axial} to form the final representation.
However, this approach is not rotationally invariant, and cannot accurately compare relative distances in non-cardinal directions.
This may not be a problem in the case of natural images, but it is an important consideration for medical images as they lack a canonical orientation and accurately capturing relative distances between biological structures may be crucial for the task.

\Textcite{wu2021rethinking} propose several relative position encodings designed specifically for images. 
The main idea is to discretise the relative Euclidean distances between input elements into a finite number of bins, and then learn a separate relative position representation for each bin.
For $k$ bins, the authors learn $k$ relative position representation vectors $\vr_1^V,\vr_2^V,\dots,\vr_k^V \in \R^{d_z}$ and then set $\vb_{ij}^K=\vr_b$ where $b$ is the index of the bin containing the relative distance $\delta_{ij}$.
The same procedure is followed for $\vb^Q_{ij}$ and $\vb^V_{ij}$.
This approach is the most similar to ours because it is rotationally invariant by incorporating Euclidean distance information into the self-attention mechanism.
However, like all of the other aforementioned relative position schemes, \textcite{wu2021rethinking} did not consider the problem of \gls{mil}.
Instead, they studied the situation where the input images were of a fixed size and split into a regular grid of tiles, so the relative distances were discrete values representing offsets between tiles.
However, in \gls{mil} there is no such restriction, and especially in the medical domain, images may be of different sizes, patches could be sampled from anywhere in the image, and they need not be sampled exhaustively.
Discretising the relative distances into bins may be problematic in \gls{mil} because the number of bins is a hyperparameter that needs to be tuned. 
If the number of bins is too small, the model cannot make distinctions between distances at the resulution required for the task at hand.
If the number of bins is too large, the model will fail to generalise to unseen distances that were not present in the training set.
In their setting, the input images were of fixed size, and the patches were exhaustively sampled from a regular grid, so the model could be trained to learn the relative position representations for all possible distances.
This is not possible in \gls{mil} because there is no restriction on the number of possible distances, and the model must generalise to unseen distance values.
Furthermore, \citeauthor{wu2021rethinking}'s model receives no explicit notion of ordering regarding the bins, so it must learn to infer this implicitly.
Our approach does not suffer from these issues, because it operates on continuous distances and explicitly encodes the ordering of the distances via the $\phi$ function in \cref{eq:bijK}.

\section{Experiments and results}
\label{sec:results}
\subsection{Toy dataset based on \smaller{MNIST}}
\label{sec:results:mnist}
Alongside their model, \textcite{ilse2018attention} introduced a toy \gls{mil} problem based on the \smaller{MNIST} dataset~\cite{deng2012mnist} which they call \smaller{MNIST-BAGS}.
In this dataset, each bag is comprised of a variable number of \smaller{MNIST} digits, \ie instances.
The bag label is positive if and only if it contains the digit ``9'' at least once.
Due to its size and simplicitly, this dataset lends itself particularly well to studying the behaviour of \gls{mil} models~\cite{li2020deep,tu2019multiple}.
However, \smaller{MNIST-BAGS} still operates under the classical \gls{mil} assumption that just one instance contributes to the bag label, so a variant of this dataset has been used where the bag label is positive if and only if it contains a pair of two specific digits, e.g. ``7'' and ``9''~\cite{li2020deep,zhao2023generalized}.

\begin{figure}
  \centering
  \begin{subfigure}[t]{.18\textwidth}
      \centering
      \includegraphics[width=\textwidth]{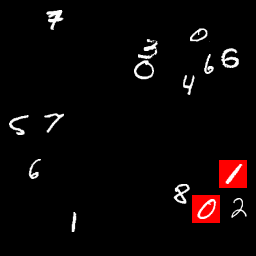}
      \caption{positive bag}
      \label{fig:mnist-collage:pos}
  \end{subfigure}\hspace{15pt}
  \begin{subfigure}[t]{.18\textwidth}
      \centering
      \includegraphics[width=\textwidth]{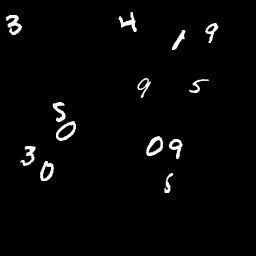}
      \caption{negative bag}
      \label{fig:mnist-collage:neg}
  \end{subfigure}\hspace{15pt}
  \begin{subfigure}[t]{.20\textwidth}
      \centering
      \resizebox{\textwidth}{!}{\import{images}{attention.pgf}}
      \caption{attention heatmap}
      \label{fig:mnist-collage:att}
  \end{subfigure}
  \caption{%
    Two bags from the \smaller{MNIST-COLLAGE} test set. 
    \textbf{(\subref{fig:mnist-collage:pos})}
    The key instances in the positive bag are in \textcolor{red}{red}.
    Note that the bag actually contains two instances of ``0'' and two instances of ``1'', but only the highlighted instances are actually close enough to each other to be considered a positive bag.
    \textbf{(\subref{fig:mnist-collage:neg})}
    In the negative bag, the ``0'' and ``1'' are too far apart for a positive label.
    Note that negative bags do not necessarily need to even contain ``0'' and ``1''.
    \textbf{(\subref{fig:mnist-collage:att})}
    Attention heatmap of DAS-MIL with key and query embeddings applied to the positive bag.
    Key instances are in \textcolor{red}{red}.
  }
  \label{fig:mnist-collage}
\end{figure}

We extend this idea to a new dataset called \smaller{MNIST-COLLAGE} that incorporates the notion of relative distances by arranging the digits in a larger image, as shown in \cref{fig:mnist-collage}.
Each bag is a collage of $n$ digits, where $n$ is sampled from a Gaussian distribution and rounded to the nearest integer.
The digits locations are sampled from a uniform distirbution encompassing the dimensions of the collage and ensuring there is no overlap. 
We label a bag positive if and only if it contains a ``0'' and a ``1'' whose centroids are within a certain Euclidean distance of each other.
Our training and test sets contain 300 and 100 bags, respectively, with the digits chosen randomly from the respective \smaller{MNIST} training and test sets, ensuring that exactly half of the bags in each set are positive.
To further increase the difficulty of the problem, we also study an inverted version of this dataset (\smaller{MNIST-COLLAGE-INV}), where the relative distance must \emph{exceed} a certain threshold.

\vspace{-8pt}\paragraph{Experimental setup}
We compare the performance of our approach to a number of existing \gls{mil} models, namely AB-MIL~\cite{ilse2018attention}, MIL-GNN~\cite{tu2019multiple}, and TransMIL~\cite{shao2021transmil}.
Additionally, we apply existing models mentioned in \cref{sec:background} that were not specifically designed for \gls{mil} to the general \gls{mil} framework outlined in \cref{sec:mil} by parameterising the aggregation function $\vg$ in \cref{eq:mil_score_embedded:embedding} with the specified model.
In this regard, we evaluate the Set Transformer~\cite{lee2019set}, a simple self-attention layer~\cite{vaswani2017attention}, as well as self-attention with various positional encodings~\cite{ramachandran2019stand,yang2021learnable}.
Importantly, we also evaluate self-attention with \citeauthor{wu2021rethinking}'s discrete relative position representations~\cite{wu2021rethinking} (with 10 encodings) as described at the end of \cref{sec:method:prior}. 

The same \gls{cnn} architecture is used for feature extraction in all models, detailed in \cref{app:mnist_collage}.
For each model, we perform a grid search over the learning rate, weight decay, and other model-sepcific hyperparameters, and report the best results across 5 runs with different seeds.
We train each model for 50 epochs with a batch size of 1 using the AdamW optimizer~\cite{loshchilov2018decoupled} on a consumer-grade CPU.
For our model, a learning rate of $10^{-3}$ and weight decay of $10^{-2}$ achieved the best results.

\begin{table}[h]
  \caption{%
    Results on the \smaller{MNIST-COLLAGE} and \smaller{MNIST-COLLAGE-INV} datasets.
    We report the mean and standard deviation of the balanced accuracy score over 5 runs (\acs{auroc} is reported in \cref{app:mnist_collage}).
    Models that contain the word ``with'' in their name parameterise the aggregation function $\vg$ in \cref{eq:mil_score_embedded:embedding} with the specified model, but these models were not originally designed for \gls{mil}.
    All models employ the same \gls{cnn} architecture for feature extraction and are trained end-to-end.
    The second column indicates the type of positional encoding used: absolute (\smaller{ABS}),  relative (\smaller{REL}) or neither (\xmark).
  }
  \label{tbl:mnist_collage}
  \adjustbox{width=\linewidth}{
    \begin{tabular}{l|c|r|rr|rr}
      \toprule
       & & & \multicolumn{2}{c|}{\smaller{MNIST-COLLAGE}} & \multicolumn{2}{c}{\smaller{MNIST-COLLAGE-INV}} \\
      Model & \multicolumn{1}{c|}{Pos} & \multicolumn{1}{c|}{Params} & \multicolumn{1}{c}{Train} & \multicolumn{1}{c|}{Test} & \multicolumn{1}{c}{Train} & \multicolumn{1}{c}{Test} \\
      \midrule
      MIL with max pool & \xmark & 15.6K & 0.846 $\pm$ 0.021 & 0.828 $\pm$ 0.033 & 0.840 $\pm$ 0.009 & 0.788 $\pm$ 0.029 \\
      AB-MIL~\cite{ilse2018attention} & \xmark & 16.1K & 0.799 $\pm$ 0.014 & 0.740 $\pm$ 0.010 & 0.805 $\pm$ 0.006 & 0.692 $\pm$ 0.015 \\
      MIL-GNN~\cite{tu2019multiple} &  \smaller{REL} & 19.2K & 0.686 $\pm$ 0.143 & 0.656 $\pm$ 0.123 & 0.784 $\pm$ 0.023 & 0.756 $\pm$ 0.019 \\
      MIL with iSet Transformer~\cite{lee2019set} & \xmark & 22.1K & 0.813 $\pm$ 0.017 & 0.734 $\pm$ 0.040 & 0.813 $\pm$ 0.023 & 0.720 $\pm$ 0.016 \\
      MIL with Set Transformer~\cite{lee2019set} & \xmark & 17.6K & 0.815 $\pm$ 0.052 & 0.832 $\pm$ 0.041 & 0.819 $\pm$ 0.018 & 0.792 $\pm$ 0.045 \\
      MIL with SA~\cite{vaswani2017attention} & \xmark & 17.2K & 0.881 $\pm$ 0.013 & 0.848 $\pm$ 0.024 & 0.840 $\pm$ 0.019 & 0.786 $\pm$ 0.077 \\
      MIL with SA + axial PE~\cite{ramachandran2019stand} &  \smaller{ABS} & 17.6K & 0.862 $\pm$ 0.008 & 0.786 $\pm$ 0.051 & 0.902 $\pm$ 0.018 & 0.774 $\pm$ 0.015 \\
      MIL with SA + Fourier PE~\cite{yang2021learnable} &  \smaller{ABS} & 18.4K & 0.878 $\pm$ 0.035 & 0.820 $\pm$ 0.022 & 0.879 $\pm$ 0.037 & 0.808 $\pm$ 0.029 \\
      MIL with disc.\ rel.\ SA~\cite{wu2021rethinking} &  \smaller{REL} & 17.8K & 0.921 $\pm$ 0.014 & 0.926 $\pm$ 0.030 & 0.877 $\pm$ 0.031 & 0.846 $\pm$ 0.060 \\
      TransMIL~\cite{shao2021transmil} &  \smaller{ABS} & 2.18M & \textbf{0.957 $\pm$ 0.029} & 0.728 $\pm$ 0.054 & \textbf{0.957 $\pm$ 0.015} & 0.742 $\pm$ 0.053 \\
      DAS-MIL (ours) &  \smaller{REL} & 17.4K & 0.955 $\pm$ 0.011 & \textbf{0.958 $\pm$ 0.013} & 0.953 $\pm$ 0.004 & \textbf{0.906 $\pm$ 0.034} \\
      \bottomrule
    \end{tabular}
  }
  \vspace{-10pt}
\end{table}

\vspace{-8pt}
\paragraph{Results}
\cref{tbl:mnist_collage} indicates that models that do not incorporate any positional information perform the worst, which is expected as the problem inherently requires the consideration of spatial correspondences.
12\% of the samples in the test set are negative bags that contain at least one ``0'' and one ``1'' which means that a model could achieve an accuracy of 88\% by ignoring the spatial information, (in effect, this task would be identical to the \smaller{MNIST-BAGS} dataset with two key instances~\cite{li2020deep,zhao2023generalized}).
While it is self-evident why the models without positional information do not exceed this score, it is important to note that none of the models that employ absolute positional encodings perform better than this baseline. 
This indicates that models with absolute positional encodings have trouble comparing relative distances between instances, which is not surprising because the absolute positional encodings are not rotationally invariant.
TransMIL~\cite{shao2021transmil} uses a type of absolute positional encoding where features are aggregated in a pyramid, and while this mechanism achieves the best training accuracy, it does not generalise well, which may be explained by the fact that their model has two orders of magnitude more parameters than all the others.
In general, models perform worse on \smaller{MNIST-COLLAGE-INV}, which is expected as long-range dependencies need to be considered.
Still, our model achieves the best test accuracy on both datasets, and is the only model that achieves a test accuracy of over 90\% on \smaller{MNIST-COLLAGE-INV}.
The discrete relative self-attention mechanism achieved the second best results, however, it uses 10 embeddings while our model uses just two.


\vspace{-8pt}\paragraph{Embeddings}
We perform an ablation study to assess how much each of the three embedding terms $\vb_{ij}^K,\vb_{ij}^Q,\vb_{ij}^V$ in \cref{eq:das:output,eq:das:compatibility} contribute to our model's performance. 
Testing all combinations of setting the embedding terms to zero, we find that our model performs worst when using just the key embeddings (\ie $\vb_{ij}^Q=\vb_{ij}^V=\vzero$), and best when using all three embeddings on \smaller{MNIST-COLLAGE}.
However, on \smaller{MNIST-COLLAGE-INV}, the model without query embeddings actually exhibits slightly better generalisation ability.
Note that setting all three embeddings to zero would reduce the model to a simple self-attention layer, which performed much worse (see Table \ref{tbl:mnist_collage}).
Furthermore, we study the effect of the term $\vb^Q_{ij} {\vb^K_{ij}}^\top$ in \cref{eq:das:compatibility} by comparing the performance of our model with and without it, and find that its subtraction improves the test accuracy by 1\% on both datasets.
As remarked in \cref{sec:distance_aware_self_attention}, this term also saves some computation in the compatibility function.
Finally, we assess whether the embedding vectors ($\vu^K,\vv^K,\vu^Q,\vv^Q,\vu^V,\vv^V$ in \cref{eq:bijK,eq:bijQV}) even need to be learned, by randomly initialising and freezing them.
Surprisingly, this decreases the mean test accuracy of our model by just 0.6\% on \smaller{MNIST-COLLAGE} and 1\% on \smaller{MNIST-COLLAGE-INV} (albeit with a larger standard deviation), indicating that the model is not very sensitive to the actual values of the embedding vectors.
However, the two trainable scalars $\beta$ and $\theta$ of the $\phi(\cdot)$ function in \cref{eq:bijK} are crucial; keeping them fixed or replacing $\phi(\cdot)$ with the identity function results in significantly worse performance.
The results of these ablation studies are summarised in \cref{tbl:ablation_embeddings} of \cref{app:mnist_collage}.

\vspace{-8pt}\paragraph{Interpreting attention maps}
We can visualise the attention weights $\alpha_{ij}$
as a heatmap to gain insight into the model's behaviour.
The attention weights are computed via the compatibility function from \cref{eq:das:compatibility} which incorporates distance information via the query and key embeddings.
However, the value embeddings are added to the output vectors from \cref{eq:das:output} after the attention weights have been computed, so their effect is not as straightforward to interpret.
Therefore, we visualise the attention weights of the version of our model that does not use value embeddings, \ie $\vb_{ij}^V=\vzero$ in the ablation study above.
\Cref{fig:mnist-collage:att} shows the attention heatmap of this model applied to the positive bag depicted in \cref{fig:mnist-collage:pos}.
This bag is interesting because it contains multiple instances of the digits ``0'' and ``1'', but only the two instances highlighted in red are close enough to be considered a positive bag.
In the attention map, the two corresponding entries are highly activated. 
By contrast, other entries in the attention matrix corresponding to the other digits that are too far apart are not activated.
Interestingly, there are two other cells that have a higher activation, corresponding to the pair of ``6'' and ``1'', which may be explained by the similarly round shape of ``6'' and ``0''.

\subsection{The \smaller{CAMELYON16} dataset}
\label{sec:results:camelyon}
The \smaller{CAMELYON16} dataset~\cite{bejnordi2017diagnostic} consists of \gls{he} stained slides of lymph node sections from breast cancer patients.
Each slide is classified either as ``tumour'' (positive) or ``normal'' (negative), which is particularly challenging because large regions in the tumour slides contain normal tissue.
We removed three slides that \textcite{liu2017detecting} identified as being labelled incorrectly, resulting in a training set of 111 tumour and 156 normal slides, and a test set of 48 tumour and 80 normal slides.

We extract patches at a resolution of $224 \times 224$ pixels from the slides in a regular grid, discarding patches that do not contain any tissue.
Due to the heterogeneity of the slides (they were acquired at two different medical centres with different scanners), we apply \citeauthor{macenko2009method}'s stain normalisation method~\cite{macenko2009method} to each patch.
The mean bag size is 4,004 in the train set, but we set the maximum bag size to 6,000 to reduce the memory requirements of the model (in larger bags, we randomly sample 6,000 patches). 
We then extract a 768-dimensional feature vector from each patch using a Transfomer model that was trained on a large dataset of \gls{he} stained patches~\cite{wang2022transformer}.
This transformer corresponds to the feature extractor $\vh$ in \cref{eq:mil_score_embedded:embedding}, with the difference that it is not trained end-to-end with the rest of the model; instead, we use the pre-trained weights from the original work~\cite{wang2022transformer} throughout to reduce the training time (they need only be computed once for the whole dataset before training).

\begin{table}[t]
  \caption{%
    Results on the \smaller{CAMELYON16} dataset.
  }
  \vspace{1pt}
  \label{tbl:camelyon16}
  \adjustbox{max width=\linewidth}{
    \begin{tabular}{l|c|c|rr|rr}
      \toprule
       & & & \multicolumn{2}{c|}{AUROC} & \multicolumn{2}{c}{Balanced accuracy} \\
      Model & Pos & \multicolumn{1}{c|}{Params} & \multicolumn{1}{c}{Train} & \multicolumn{1}{c|}{Test} & \multicolumn{1}{c}{Train} & \multicolumn{1}{c}{Test} \\
      \midrule
      MIL with max pool & \xmark & 769 & 0.686 $\pm$ 0.018 & 0.739 $\pm$ 0.017 & 0.639 $\pm$ 0.033 & 0.573 $\pm$ 0.077 \\
      AB-MIL~\cite{ilse2018attention} & \xmark & 8.47K & 0.964 $\pm$ 0.003 & 0.795 $\pm$ 0.009 & 0.923 $\pm$ 0.004 & 0.773 $\pm$ 0.010 \\
      MIL with SA~\cite{vaswani2017attention} & \xmark & 27.7K & 0.972 $\pm$ 0.014 & 0.823 $\pm$ 0.050 & 0.925 $\pm$ 0.016 & 0.803 $\pm$ 0.033 \\
      MIL with SA + axial PE~\cite{ramachandran2019stand} &  \smaller{ABS} & 27.7K & 0.956 $\pm$ 0.016 & 0.485 $\pm$ 0.018 & 0.887 $\pm$ 0.017 & 0.490 $\pm$ 0.034 \\
      MIL with SA + Fourier PE~\cite{yang2021learnable} &  \smaller{ABS} & 41K & 0.975 $\pm$ 0.007 & 0.819 $\pm$ 0.028 & 0.932 $\pm$ 0.011 & 0.770 $\pm$ 0.041 \\
      MIL with disc.\ rel.\ SA~\cite{wu2021rethinking} &  \smaller{REL} & 28K & 0.976 $\pm$ 0.006 & 0.806 $\pm$ 0.015 & 0.926 $\pm$ 0.017 & 0.787 $\pm$ 0.008 \\
      TransMIL~\cite{shao2021transmil} &  \smaller{ABS} & 2.54M & \textbf{0.999 $\pm$ 0.002} & 0.911 $\pm$ 0.027 & \textbf{0.997 $\pm$ 0.007} & 0.857 $\pm$ 0.032 \\
      DAS-MIL (ours) &  \smaller{REL} & 412K & 0.997 $\pm$ 0.005 & \textbf{0.914 $\pm$ 0.007} & 0.978 $\pm$ 0.022 & \textbf{0.864 $\pm$ 0.025} \\
      \bottomrule
    \end{tabular}
  }
  \vspace{-15pt}
\end{table}

Like TransMIL~\cite{shao2021transmil}, we add a trainable fully connected layer with ReLU activation at the start of our model to reduce the feature size from 768 to $d_z=512$.
To find the best hyperparameters for every model, we perform a grid search over the learning rate, weight decay, and optimizer (AdamW~\cite{loshchilov2018decoupled} or Lookahead~\cite{zhang2019lookahead}), training every model for 30 epochs and reporting the best results in \cref{tbl:camelyon16}.
We find the best results with the Lookahead optimizer~\cite{zhang2019lookahead} in the tested models.
On a single GPU, our model takes approximately three hours to train which is comparable to TransMIL~\cite{shao2021transmil} and the \gls{mil} variant that uses self-attention with discrete relative position representations~\cite{wu2021rethinking}.
The other models such as AB-MIL~\cite{ilse2018attention} and vanilla self-attention are much simpler and require about half that time, but do not achieve competitive results.
For TransMIL~\cite{shao2021transmil}, which in \citeauthor{shao2021transmil}'s original work also uses the Lookahead optimizer, our hyperparameter search yielded the same optimal learning rate and weight decay as in their work, but we were unable to reproduce their reported results on \smaller{CAMELYON16} (we achieve a test \gls{auroc} of 0.911 with their model compared to their reported 0.909).
This may be due to differences in the preprocessing steps or due to the random seed (we report mean and standard deviation across 5 runs with different random seeds, but \citeauthor{shao2021transmil} do not specify how many runs they performed or include the standard deviation in their results).
While TransMIL achieves better \gls{auroc} and balanced accuracy scores on the training set, our model generalises better to the test set, where our model outperforms all other approaches.

\section{Conclusion}
\label{sec:conclusion}

Some image-based \gls{mil} problems, especially in the domain of medical imaging, require the model to consider the spatial arrangement of the tiles for prediction.
Recognising this fact, we develop a self-attention-based \gls{mil} model that is able to compare relative distances using a novel scheme for relative distance representations. 
We validate our model on two datasets: a custom dataset designed to test the ability to compare relative distances, and \smaller{CAMELYON16}.
We have open-sourced the model's code to ensure that our model can be employed and validated on more datasets.

\vspace{-8pt}\paragraph{Limitations and future work}
A limitation of our model is that it is interpretable only in certain configurations (\ie without value embeddings).
We leave improving the model's interpretability for future work.
Furthermore, it would be interesting to study variations of the interpolation function $\phi(x)$, and whether it could be used to incorporate inductive biases.

\vspace{-8pt}\paragraph{Broader impact}
We hope our models inspire the development of more relative distance-based approaches.
As mentioned in the introduction, researchers in computational pathology have shown that spatial information is crucial for forming diagnoses in various problems, so relative distance-based approaches have the potential to improve diagnostic quality.
In that vein, we would also encourage more research in the interpretability and safety of these models. 

\printbibliography

@article{dietterich1997solving,
  title     = {Solving the multiple instance problem with axis-parallel rectangles},
  author    = {Dietterich, Thomas G and Lathrop, Richard H and Lozano-P{\'e}rez, Tom{\'a}s},
  journal   = {Artificial intelligence},
  volume    = {89},
  number    = {1-2},
  pages     = {31--71},
  year      = {1997},
  publisher = {Elsevier}
}

@inproceedings{maron1997framework,
  author    = {Maron, Oded and Lozano-P\'{e}rez, Tom\'{a}s},
  booktitle = {Advances in Neural Information Processing Systems},
  title     = {A Framework for Multiple-Instance Learning},
  volume    = {10},
  year      = {1997}
}

@inproceedings{ilse2018attention,
  title        = {Attention-based deep multiple instance learning},
  author       = {Ilse, Maximilian and Tomczak, Jakub and Welling, Max},
  booktitle    = {International conference on machine learning},
  pages        = {2127--2136},
  year         = {2018},
  organization = {PMLR}
}

@inproceedings{lee2019set,
  title        = {Set transformer: A framework for attention-based permutation-invariant neural networks},
  author       = {Lee, Juho and Lee, Yoonho and Kim, Jungtaek and Kosiorek, Adam and Choi, Seungjin and Teh, Yee Whye},
  booktitle    = {International conference on machine learning},
  pages        = {3744--3753},
  year         = {2019},
  organization = {PMLR}
}

@inproceedings{shaw2018self,
  title     = {Self-Attention with Relative Position Representations},
  author    = {Peter Shaw and Jakob Uszkoreit and Ashish Vaswani},
  booktitle = {North American Chapter of the Association for Computational Linguistics},
  year      = {2018}
}

@inproceedings{ying2021transformers,
  author    = {Ying, Chengxuan and Cai, Tianle and Luo, Shengjie and Zheng, Shuxin and Ke, Guolin and He, Di and Shen, Yanming and Liu, Tie-Yan},
  booktitle = {Advances in Neural Information Processing Systems},
  pages     = {28877--28888},
  title     = {Do Transformers Really Perform Badly for Graph Representation?},
  volume    = {34},
  year      = {2021}
}

@article{campanella2019clinical,
  title     = {Clinical-grade computational pathology using weakly supervised deep learning on whole slide images},
  author    = {Campanella, Gabriele and Hanna, Matthew G and Geneslaw, Luke and Miraflor, Allen and Werneck Krauss Silva, Vitor and Busam, Klaus J and Brogi, Edi and Reuter, Victor E and Klimstra, David S and Fuchs, Thomas J},
  journal   = {Nature medicine},
  volume    = {25},
  number    = {8},
  pages     = {1301--1309},
  year      = {2019},
  publisher = {Nature Publishing Group US New York}
}

@article{niehues2023generalisable,
  title    = {Generalizable biomarker prediction from cancer pathology slides with self-supervised deep learning: A retrospective multi-centric study},
  journal  = {Cell Reports Medicine},
  pages    = {100980},
  year     = {2023},
  issn     = {2666-3791},
  author   = {Jan Moritz Niehues and Philip Quirke and Nicholas P. West and Heike I. Grabsch and Marko {van Treeck} and Yoni Schirris and Gregory P. Veldhuizen and Gordon G.A. Hutchins and Susan D. Richman and Sebastian Foersch and Titus J. Brinker and Junya Fukuoka and Andrey Bychkov and Wataru Uegami and Daniel Truhn and Hermann Brenner and Alexander Brobeil and Michael Hoffmeister and Jakob Nikolas Kather}
}

@article{saldanha2023self,
  title     = {Self-supervised attention-based deep learning for pan-cancer mutation prediction from histopathology},
  author    = {Saldanha, Oliver Lester and Loeffler, Chiara ML and Niehues, Jan Moritz and van Treeck, Marko and Seraphin, Tobias P and Hewitt, Katherine Jane and Cifci, Didem and Veldhuizen, Gregory Patrick and Ramesh, Siddhi and Pearson, Alexander T and others},
  journal   = {NPJ Precision Oncology},
  volume    = {7},
  number    = {1},
  pages     = {35},
  year      = {2023},
  publisher = {Nature Publishing Group UK London}
}

@article{wang2018revisiting,
  title     = {Revisiting multiple instance neural networks},
  author    = {Wang, Xinggang and Yan, Yongluan and Tang, Peng and Bai, Xiang and Liu, Wenyu},
  journal   = {Pattern Recognition},
  volume    = {74},
  pages     = {15--24},
  year      = {2018},
  publisher = {Elsevier}
}

@inproceedings{rymarczyk2021kernel,
  title     = {Kernel self-attention for weakly-supervised image classification using deep multiple instance learning},
  author    = {Rymarczyk, Dawid and Borowa, Adriana and Tabor, Jacek and Zielinski, Bartosz},
  booktitle = {Proceedings of the IEEE/CVF Winter Conference on Applications of Computer Vision},
  pages     = {1721--1730},
  year      = {2021}
}

@article{shao2021transmil,
  title   = {Transmil: Transformer based correlated multiple instance learning for whole slide image classification},
  author  = {Shao, Zhuchen and Bian, Hao and Chen, Yang and Wang, Yifeng and Zhang, Jian and Ji, Xiangyang and others},
  journal = {Advances in neural information processing systems},
  volume  = {34},
  pages   = {2136--2147},
  year    = {2021}
}

@article{vaswani2017attention,
  title   = {Attention is all you need},
  author  = {Vaswani, Ashish and Shazeer, Noam and Parmar, Niki and Uszkoreit, Jakob and Jones, Llion and Gomez, Aidan N and Kaiser, {\L}ukasz and Polosukhin, Illia},
  journal = {Advances in neural information processing systems},
  volume  = {30},
  year    = {2017}
}

@inproceedings{fu2012implementation,
  title        = {Implementation of multiple-instance learning in drug activity prediction},
  author       = {Fu, Gang and Nan, Xiaofei and Liu, Haining and Patel, Ronak Y and Daga, Pankaj R and Chen, Yixin and Wilkins, Dawn E and Doerksen, Robert J},
  booktitle    = {BMC bioinformatics},
  volume       = {13},
  number       = {15},
  pages        = {1--12},
  year         = {2012},
  organization = {BioMed Central}
}

@inproceedings{pappas2014explaining,
  title     = {Explaining the stars: Weighted multiple-instance learning for aspect-based sentiment analysis},
  author    = {Pappas, Nikolaos and Popescu-Belis, Andrei},
  booktitle = {Proceedings of the 2014 Conference on Empirical Methods In Natural Language Processing (EMNLP)},
  pages     = {455--466},
  year      = {2014}
}

@inproceedings{wan2019c,
  title     = {C-mil: Continuation multiple instance learning for weakly supervised object detection},
  author    = {Wan, Fang and Liu, Chang and Ke, Wei and Ji, Xiangyang and Jiao, Jianbin and Ye, Qixiang},
  booktitle = {Proceedings of the IEEE/CVF Conference on Computer Vision and Pattern Recognition},
  pages     = {2199--2208},
  year      = {2019}
}

@inproceedings{wu2015deep,
  title     = {Deep multiple instance learning for image classification and auto-annotation},
  author    = {Wu, Jiajun and Yu, Yinan and Huang, Chang and Yu, Kai},
  booktitle = {Proceedings of the IEEE conference on computer vision and pattern recognition},
  pages     = {3460--3469},
  year      = {2015}
}

@inproceedings{yang2021learnable,
  author    = {Li, Yang and Si, Si and Li, Gang and Hsieh, Cho-Jui and Bengio, Samy},
  booktitle = {Advances in Neural Information Processing Systems},
  pages     = {15816--15829},
  title     = {Learnable Fourier Features for Multi-dimensional Spatial Positional Encoding},
  volume    = {34},
  year      = {2021}
}

@inproceedings{carion2020end,
  title     = {End-to-end object detection with transformers},
  author    = {Carion, Nicolas and Massa, Francisco and Synnaeve, Gabriel and Usunier, Nicolas and Kirillov, Alexander and Zagoruyko, Sergey},
  booktitle = {Computer Vision--ECCV 2020: 16th European Conference, Glasgow, UK, August 23--28, 2020, Proceedings, Part I 16},
  pages     = {213--229},
  year      = {2020}
}

@inproceedings{dosovitskiy2021image,
  title     = {An Image is Worth 16x16 Words: Transformers for Image Recognition at Scale},
  author    = {Alexey Dosovitskiy and Lucas Beyer and Alexander Kolesnikov and Dirk Weissenborn and Xiaohua Zhai and Thomas Unterthiner and Mostafa Dehghani and Matthias Minderer and Georg Heigold and Sylvain Gelly and Jakob Uszkoreit and Neil Houlsby},
  booktitle = {International Conference on Learning Representations},
  year      = {2021}
}

@inproceedings{parmar2018image,
  title     = {Image transformer},
  author    = {Parmar, Niki and Vaswani, Ashish and Uszkoreit, Jakob and Kaiser, Lukasz and Shazeer, Noam and Ku, Alexander and Tran, Dustin},
  booktitle = {International conference on machine learning},
  pages     = {4055--4064},
  year      = {2018}
}

@article{wagner2023fully,
  title   = {Fully transformer-based biomarker prediction from colorectal cancer histology: a large-scale multicentric study},
  author  = {Sophia J. Wagner and Daniel Reisenbüchler and Nicholas P. West and Jan Moritz Niehues and Gregory Patrick Veldhuizen and Philip Quirke and Heike I. Grabsch and Piet A. van den Brandt and Gordon G. A. Hutchins and Susan D. Richman and Tanwei Yuan and Rupert Langer and Josien Christina Anna Jenniskens and Kelly Offermans and Wolfram Mueller and Richard Gray and Stephen B. Gruber and Joel K. Greenson and Gad Rennert and Joseph D. Bonner and Daniel Schmolze and Jacqueline A. James and Maurice B. Loughrey and Manuel Salto-Tellez and Hermann Brenner and Michael Hoffmeister and Daniel Truhn and Julia A. Schnabel and Melanie Boxberg and Tingying Peng and Jakob Nikolas Kather},
  journal = {arXiv preprint arXiv:2301.09617},
  year    = {2023}
}

@inproceedings{li2020deep,
  title     = {Deep multi-instance learning with induced self-attention for medical image classification},
  author    = {Li, Zhenliang and Yuan, Liming and Xu, Haixia and Cheng, Rui and Wen, Xianbin},
  booktitle = {2020 IEEE International Conference on Bioinformatics and Biomedicine (BIBM)},
  pages     = {446--450},
  year      = {2020}
}

@article{tu2019multiple,
  title   = {Multiple instance learning with graph neural networks},
  author  = {Tu, Ming and Huang, Jing and He, Xiaodong and Zhou, Bowen},
  journal = {arXiv preprint arXiv:1906.04881},
  year    = {2019}
}

@article{zhao2023generalized,
  title   = {Generalized attention-based deep multi-instance learning},
  author  = {Zhao, Lu and Yuan, Liming and Hao, Kun and Wen, Xianbin},
  journal = {Multimedia Systems},
  volume  = {29},
  number  = {1},
  pages   = {275--287},
  year    = {2023}
}

@article{xiong2023diagnose,
  title   = {Diagnose Like a Pathologist: Transformer-Enabled Hierarchical Attention-Guided Multiple Instance Learning for Whole Slide Image Classification},
  author  = {Xiong, Conghao and Chen, Hao and Sung, Joseph and King, Irwin},
  journal = {arXiv preprint arXiv:2301.08125},
  year    = {2023}
}

@inproceedings{huang2021integration,
  title     = {Integration of patch features through self-supervised learning and transformer for survival analysis on whole slide images},
  author    = {Huang, Ziwang and Chai, Hua and Wang, Ruoqi and Wang, Haitao and Yang, Yuedong and Wu, Hejun},
  booktitle = {Medical Image Computing and Computer Assisted Intervention--MICCAI 2021: 24th International Conference, Strasbourg, France, September 27--October 1, 2021, Proceedings, Part VIII 24},
  pages     = {561--570},
  year      = {2021}
}

@inproceedings{chen2021multimodal,
  title     = {Multimodal co-attention transformer for survival prediction in gigapixel whole slide images},
  author    = {Chen, Richard J and Lu, Ming Y and Weng, Wei-Hung and Chen, Tiffany Y and Williamson, Drew FK and Manz, Trevor and Shady, Maha and Mahmood, Faisal},
  booktitle = {Proceedings of the IEEE/CVF International Conference on Computer Vision},
  pages     = {4015--4025},
  year      = {2021}
}

@inproceedings{myronenko2021accounting,
  title     = {Accounting for dependencies in deep learning based multiple instance learning for whole slide imaging},
  author    = {Myronenko, Andriy and Xu, Ziyue and Yang, Dong and Roth, Holger R and Xu, Daguang},
  booktitle = {Medical Image Computing and Computer Assisted Intervention--MICCAI 2021: 24th International Conference, Strasbourg, France, September 27--October 1, 2021, Proceedings, Part VIII 24},
  pages     = {329--338},
  year      = {2021}
}

@inproceedings{chen2022scaling,
  title     = {Scaling vision transformers to gigapixel images via hierarchical self-supervised learning},
  author    = {Chen, Richard J and Chen, Chengkuan and Li, Yicong and Chen, Tiffany Y and Trister, Andrew D and Krishnan, Rahul G and Mahmood, Faisal},
  booktitle = {Proceedings of the IEEE/CVF Conference on Computer Vision and Pattern Recognition},
  pages     = {16144--16155},
  year      = {2022}
}

@inproceedings{loshchilov2018decoupled,
  title     = {Decoupled Weight Decay Regularization},
  author    = {Ilya Loshchilov and Frank Hutter},
  booktitle = {International Conference on Learning Representations},
  year      = {2019}
}

@inproceedings{wu2021rethinking,
  title     = {Rethinking and improving relative position encoding for vision transformer},
  author    = {Wu, Kan and Peng, Houwen and Chen, Minghao and Fu, Jianlong and Chao, Hongyang},
  booktitle = {Proceedings of the IEEE/CVF International Conference on Computer Vision},
  pages     = {10033--10041},
  year      = {2021}
}

@inproceedings{liutkus2021relative,
  title        = {Relative positional encoding for transformers with linear complexity},
  author       = {Liutkus, Antoine and C{\i}́fka, Ond{\v{r}}ej and Wu, Shih-Lun and Simsekli, Umut and Yang, Yi-Hsuan and Richard, Gael},
  booktitle    = {International Conference on Machine Learning},
  pages        = {7067--7079},
  year         = {2021},
  organization = {PMLR}
}

@inproceedings{qian2022transformer,
  title        = {Transformer based multiple instance learning for weakly supervised histopathology image segmentation},
  author       = {Qian, Ziniu and Li, Kailu and Lai, Maode and Chang, Eric I-Chao and Wei, Bingzheng and Fan, Yubo and Xu, Yan},
  booktitle    = {Medical Image Computing and Computer Assisted Intervention--MICCAI 2022: 25th International Conference, Singapore, September 18--22, 2022, Proceedings, Part II},
  pages        = {160--170},
  year         = {2022},
  organization = {Springer}
}

@article{li2020dual,
  title   = {Dual-stream maximum self-attention multi-instance learning},
  author  = {Li, Bin and Eliceiri, Kevin W},
  journal = {arXiv preprint arXiv:2006.05538},
  year    = {2020}
}

@inproceedings{ding2022deep,
  title        = {Deep Multi-Instance Learning with Adaptive Recurrent Pooling for Medical Image Classification},
  author       = {Ding, Yi and Zhao, Lu and Yuan, Liming and Wen, Xianbin},
  booktitle    = {2022 IEEE International Conference on Bioinformatics and Biomedicine (BIBM)},
  pages        = {3335--3342},
  year         = {2022},
  organization = {IEEE}
}

@inproceedings{dai2019transformer,
  title     = {Transformer-{XL}: Attentive Language Models beyond a Fixed-Length Context},
  author    = {Dai, Zihang  and
               Yang, Zhilin  and
               Yang, Yiming  and
               Carbonell, Jaime  and
               Le, Quoc  and
               Salakhutdinov, Ruslan},
  booktitle = {Proceedings of the 57th Annual Meeting of the Association for Computational Linguistics},
  month     = {6},
  year      = {2019},
  address   = {Florence, Italy},
  publisher = {Association for Computational Linguistics},
  pages     = {2978--2988}
}

@inproceedings{huang2020improve,
  title     = {Improve Transformer Models with Better Relative Position Embeddings},
  author    = {Huang, Zhiheng  and
               Liang, Davis  and
               Xu, Peng  and
               Xiang, Bing},
  booktitle = {Findings of the Association for Computational Linguistics: EMNLP 2020},
  month     = {11},
  year      = {2020},
  publisher = {Association for Computational Linguistics},
  pages     = {3327--3335}
}

@inproceedings{ramachandran2019stand,
  author    = {Ramachandran, Prajit and Parmar, Niki and Vaswani, Ashish and Bello, Irwan and Levskaya, Anselm and Shlens, Jon},
  booktitle = {Advances in Neural Information Processing Systems},
  editor    = {H. Wallach and H. Larochelle and A. Beygelzimer and F. d\textquotesingle Alch\'{e}-Buc and E. Fox and R. Garnett},
  pages     = {},
  publisher = {Curran Associates, Inc.},
  title     = {Stand-Alone Self-Attention in Vision Models},
  volume    = {32},
  year      = {2019}
}

@inproceedings{wang2020axial,
  title     = {Axial-DeepLab: Stand-Alone Axial-Attention for Panoptic Segmentation},
  author    = {Wang, Huiyu and Zhu, Yukun and Green, Bradley and Adam, Hartwig and Yuille, Alan and Chen, Liang-Chieh},
  booktitle = {European Conference on Computer Vision (ECCV)},
  year      = {2020}
}

@article{deng2012mnist,
  title     = {The mnist database of handwritten digit images for machine learning research [best of the web]},
  author    = {Deng, Li},
  journal   = {IEEE signal processing magazine},
  volume    = {29},
  number    = {6},
  pages     = {141--142},
  year      = {2012},
  publisher = {IEEE}
}

@article{elnahhas2023regression,
  title   = {Regression-based Deep-Learning predicts molecular biomarkers from pathology slides},
  author  = {El Nahhas, Omar SM and Loeffler, Chiara ML and Carrero, Zunamys I and van Treeck, Marko and Kolbinger, Fiona R and Hewitt, Katherine J and Muti, Hannah S and Graziani, Mara and Zeng, Qinghe and Calderaro, Julien and others},
  journal = {arXiv preprint arXiv:2304.05153},
  year    = {2023}
}

@article{nearchou2019automated,
  title     = {Automated Analysis of Lymphocytic Infiltration, Tumor Budding, and Their Spatial Relationship Improves Prognostic Accuracy in Colorectal CancerAutomated Image Analysis in Colorectal Cancer Prognosis},
  author    = {Nearchou, Ines P and Lillard, Kate and Gavriel, Christos G and Ueno, Hideki and Harrison, David J and Caie, Peter D},
  journal   = {Cancer immunology research},
  volume    = {7},
  number    = {4},
  pages     = {609--620},
  year      = {2019},
  publisher = {AACR}
}

@article{todd2019identifying,
  author  = {Jones-Todd, Charlotte M. and Caie, Peter and Illian, Janine B. and Stevenson, Ben C. and Savage, Anne and Harrison, David J. and Bown, James L.},
  title   = {Identifying prognostic structural features in tissue sections of colon cancer patients using point pattern analysis},
  journal = {Statistics in Medicine},
  volume  = {38},
  number  = {8},
  pages   = {1421-1441},
  year    = {2019}
}

@article{bejnordi2017diagnostic,
  title     = {Diagnostic assessment of deep learning algorithms for detection of lymph node metastases in women with breast cancer},
  author    = {Bejnordi, Babak Ehteshami and Veta, Mitko and Van Diest, Paul Johannes and Van Ginneken, Bram and Karssemeijer, Nico and Litjens, Geert and Van Der Laak, Jeroen AWM and Hermsen, Meyke and Manson, Quirine F and Balkenhol, Maschenka and others},
  journal   = {JAMA},
  volume    = {318},
  number    = {22},
  pages     = {2199--2210},
  year      = {2017},
  publisher = {American Medical Association}
}

@article{defilippis2022use,
  title     = {Use of high-plex data reveals novel insights into the tumour microenvironment of clear cell renal cell carcinoma},
  author    = {De Filippis, Raffaele and W{\"o}lflein, Georg and Um, In Hwa and Caie, Peter D and Warren, Sarah and White, Andrew and Suen, Elizabeth and To, Emily and Arandjelovi{\'c}, Ognjen and Harrison, David J},
  journal   = {Cancers},
  volume    = {14},
  number    = {21},
  pages     = {5387},
  year      = {2022},
  publisher = {MDPI}
}

@article{hackl2016computational,
  title     = {Computational genomics tools for dissecting tumour--immune cell interactions},
  author    = {Hackl, Hubert and Charoentong, Pornpimol and Finotello, Francesca and Trajanoski, Zlatko},
  journal   = {Nature Reviews Genetics},
  volume    = {17},
  number    = {8},
  pages     = {441--458},
  year      = {2016},
  publisher = {Nature Publishing Group UK London}
}

@article{shelton2021engineering,
  title     = {Engineering approaches for studying immune-tumor cell interactions and immunotherapy},
  author    = {Shelton, Sarah E and Nguyen, Huu Tuan and Barbie, David A and Kamm, Roger D},
  journal   = {Iscience},
  volume    = {24},
  number    = {1},
  pages     = {101985},
  year      = {2021},
  publisher = {Elsevier}
}

@inproceedings{velickovic2018graph,
  title     = {Graph Attention Networks},
  author    = {Veli{\v{c}}kovi{\'{c}}, Petar and Cucurull, Guillem and Casanova, Arantxa and Romero, Adriana and Li{\`{o}}, Pietro and Bengio, Yoshua},
  booktitle = {International Conference on Learning Representations},
  year      = {2018}
}

@inproceedings{kipf2017semisupervised,
  title     = {Semi-Supervised Classification with Graph Convolutional Networks},
  author    = {Thomas N. Kipf and Max Welling},
  booktitle = {International Conference on Learning Representations},
  year      = {2017}
}

@article{liu2017detecting,
  title   = {Detecting cancer metastases on gigapixel pathology images},
  author  = {Liu, Yun and Gadepalli, Krishna and Norouzi, Mohammad and Dahl, George E and Kohlberger, Timo and Boyko, Aleksey and Venugopalan, Subhashini and Timofeev, Aleksei and Nelson, Philip Q and Corrado, Greg S and others},
  journal = {arXiv preprint arXiv:1703.02442},
  year    = {2017}
}

@inproceedings{macenko2009method,
  title     = {A method for normalizing histology slides for quantitative analysis},
  author    = {Macenko, Marc and Niethammer, Marc and Marron, James S and Borland, David and Woosley, John T and Guan, Xiaojun and Schmitt, Charles and Thomas, Nancy E},
  booktitle = {IEEE international symposium on biomedical imaging},
  pages     = {1107--1110},
  year      = {2009}
}

@article{wang2022transformer,
  title     = {Transformer-based unsupervised contrastive learning for histopathological image classification},
  author    = {Wang, Xiyue and Yang, Sen and Zhang, Jun and Wang, Minghui and Zhang, Jing and Yang, Wei and Huang, Junzhou and Han, Xiao},
  journal   = {Medical Image Analysis},
  volume    = {81},
  pages     = {102559},
  year      = {2022},
  publisher = {Elsevier}
}

@article{zhang2019lookahead,
  title   = {Lookahead optimizer: k steps forward, 1 step back},
  author  = {Zhang, Michael and Lucas, James and Ba, Jimmy and Hinton, Geoffrey E},
  journal = {Advances in neural information processing systems},
  volume  = {32},
  year    = {2019}
}

\clearpage
\appendix
\section{Details of the \smaller{MNIST-COLLAGE} experiments}
\label{app:mnist_collage}

In the experiments involving the \smaller{MNIST-COLLAGE} and \smaller{MNIST-COLLAGE-INV} datasets, we use a \gls{cnn} as the feature extractor $\vh$ in \cref{eq:mil_score_embedded:embedding}.
Each model is trained end-to-end, \ie the feature extractor is trained alongside the rest of the model, and they all use the same \gls{cnn} architecture, described as follows:
a convolutional layer (10 output channels, kernel size 5, stride 1), ReLU activation, max pooling (kernel size 2, stride 2), dropout (probability 0.1);
another convolutional layer (20 output channels, kernel size 5, stride 1), ReLU activation, max pooling (kernel size 2, stride 2);
the flattened outputs are then passed through a dropout layer (probability 0.5) and a linear layer (32 output units) activated by ReLU.
This means that we employ a feature size $d_z=32$ using the notation from \cref{sec:mil}.

\paragraph{Hyperparameters}
The best hyperparameters found in the grid search are shown in \cref{tbl:mnist_collage_hyperparams} for each model on the \smaller{MNIST-COLLAGE} dataset.
One of the hyperparameters we considered was the final part of the feature aggregation function $\vg$ which in \cref{eq:method:z_prime} is the maximum function.
We find that every model's best hyperparameter configuration uses max instead of mean pooling, including the \glspl{gnn}, where the tested aggregation functions were global mean and global max pooling.
It makes sense that max pooling is more suitable for \gls{mil} because just a small number of key instances are responsible for the bag label, so the corresponding features may be diluted by mean pooling.

\begin{table}[h]
  \caption{Best hyperparameters found via grid search for each model on the \smaller{MNIST-COLLAGE} dataset.}
  \label{tbl:mnist_collage_hyperparams}
  \adjustbox{max width=\linewidth}{
    \begin{tabular}{l|rrrrr}
      \toprule
      Model & optimiser & LR & weight decay & hidden dim & agg \\
      \midrule
      MIL with max pool & Adam & 0.0001 & 0.01 & 10 & max \\
      AB-MIL~\cite{ilse2018attention} & Adam & 0.0001 & 0.001 & 15 & N/A \\
      MIL with GNN (GAT~\cite{velickovic2018graph}) & Adam & 0.001 & 0.1 & 20 & max \\
      MIL with GNN (GCN~\cite{kipf2017semisupervised}) & Adam & 0.001 & 0.1 & 15 & max \\
      MIL-GNN~\cite{tu2019multiple} & Adam & 0.001 & 0.01 & 20 & N/A \\
      MIL with iSet Transformer~\cite{lee2019set} & Adam & 0.0001 & 0.1 & 15 & N/A \\
      MIL with Set Transformer~\cite{lee2019set} & Adam & 0.001 & 0.1 & 10 & N/A \\
      MIL with SA~\cite{vaswani2017attention} & Adam & 0.001 & 0.1 & 10 & max \\
      MIL with SA + axial PE~\cite{ramachandran2019stand} & Adam & 0.0001 & 0.1 & 15 & max \\
      MIL with SA + Fourier PE~\cite{yang2021learnable} & Adam & 0.001 & 0.001 & 15 & max \\
      MIL with disc.\ rel.\ SA~\cite{wu2021rethinking} & Adam & 0.0001 & 0.1 & 10 & max \\
      TransMIL~\cite{shao2021transmil} & Lookahead & 0.0001 & 0.1 & N/A & N/A \\
      DAS-MIL (ours) & Adam & 0.001 & 0.01 & 10 & max \\
      \bottomrule
    \end{tabular}
  }
\end{table}

\paragraph{Further results}
For completeness, we report the results of experiments we conducted using \gls{mil} with two different types of \glspl{gnn} (employing GAT~\cite{velickovic2018graph} or GCN~\cite{kipf2017semisupervised} layers) in \cref{tbl:mnist_collage:further_results}.
We did not include these results in the main paper to avoid cluttering \cref{tbl:mnist_collage}.
We also report the \gls{auroc} scores of the \smaller{MNIST-COLLAGE} and \smaller{MNIST-COLLAGE-INV} experiments in \cref{tbl:mnist_collage:auroc}.
\begin{table}[h]
  \caption{%
    Further results on the \smaller{MNIST-COLLAGE} and \smaller{MNIST-COLLAGE-INV} datasets (reporting balanced accuracy scores).}
  \label{tbl:mnist_collage:further_results}
  \adjustbox{width=\linewidth}{
    \begin{tabular}{l|c|r|rr|rr}
      \toprule
       & & & \multicolumn{2}{c|}{\smaller{MNIST-COLLAGE}} & \multicolumn{2}{c}{\smaller{MNIST-COLLAGE-INV}} \\
      Model & \multicolumn{1}{c|}{Pos} & \multicolumn{1}{c|}{Params} & \multicolumn{1}{c}{Train} & \multicolumn{1}{c|}{Test} & \multicolumn{1}{c}{Train} & \multicolumn{1}{c}{Test} \\
      \midrule
      MIL with GNN (GAT~\cite{velickovic2018graph}) &  \smaller{REL} & 16.7K & 0.811 $\pm$ 0.089 & 0.758 $\pm$ 0.041 & 0.745 $\pm$ 0.033 & 0.716 $\pm$ 0.018 \\
      MIL with GNN (GCN~\cite{kipf2017semisupervised}) &  \smaller{REL} & 16.3K & 0.865 $\pm$ 0.031 & 0.790 $\pm$ 0.058 & 0.883 $\pm$ 0.034 & 0.794 $\pm$ 0.036 \\
      \bottomrule
    \end{tabular}
  }
  \vspace{-10pt}
\end{table}

\begin{table}[h]
  \caption{\Acs{auroc} scores on the \smaller{MNIST-COLLAGE} and \smaller{MNIST-COLLAGE-INV} datasets.}
  \label{tbl:mnist_collage:auroc}
  \adjustbox{max width=\linewidth}{
    \begin{tabular}{l|c|r|rr|rr}
      \toprule
      & & & \multicolumn{2}{c|}{\smaller{MNIST-COLLAGE}} & \multicolumn{2}{c}{\smaller{MNIST-COLLAGE-INV}} \\
      Model & \multicolumn{1}{c|}{Pos} & \multicolumn{1}{c|}{Params} & \multicolumn{1}{c}{Train} & \multicolumn{1}{c|}{Test} & \multicolumn{1}{c}{Train} & \multicolumn{1}{c}{Test} \\
      \midrule
      MIL with max pool & \xmark & 15.6K & 0.924 $\pm$ 0.007 & 0.905 $\pm$ 0.008 & 0.912 $\pm$ 0.008 & 0.830 $\pm$ 0.010 \\
      AB-MIL~\cite{ilse2018attention} & \xmark & 16.1K & 0.893 $\pm$ 0.017 & 0.825 $\pm$ 0.008 & 0.890 $\pm$ 0.012 & 0.790 $\pm$ 0.008 \\
      MIL with GNN (GAT~\cite{velickovic2018graph}) &  \smaller{REL} & 16.7K & 0.873 $\pm$ 0.086 & 0.820 $\pm$ 0.068 & 0.795 $\pm$ 0.045 & 0.739 $\pm$ 0.032 \\
      MIL with GNN (GCN~\cite{kipf2017semisupervised}) &  \smaller{REL} & 16.3K & 0.932 $\pm$ 0.029 & 0.851 $\pm$ 0.033 & 0.941 $\pm$ 0.027 & 0.886 $\pm$ 0.013 \\
      MIL with iSet Transformer~\cite{lee2019set} & \xmark & 22.1K & 0.889 $\pm$ 0.008 & 0.829 $\pm$ 0.013 & 0.881 $\pm$ 0.011 & 0.808 $\pm$ 0.004 \\
      MIL with Set Transformer~\cite{lee2019set} & \xmark & 17.6K & 0.869 $\pm$ 0.064 & 0.867 $\pm$ 0.036 & 0.872 $\pm$ 0.014 & 0.855 $\pm$ 0.024 \\
      MIL-GNN~\cite{tu2019multiple} &  \smaller{REL} & 19.2K & 0.648 $\pm$ 0.022 & 0.708 $\pm$ 0.059 & 0.723 $\pm$ 0.056 & 0.716 $\pm$ 0.071 \\
      MIL-GNN-DS~\cite{tu2019multiple} &  \smaller{REL} & 19.2K & 0.771 $\pm$ 0.130 & 0.778 $\pm$ 0.093 & 0.864 $\pm$ 0.027 & 0.817 $\pm$ 0.008 \\
      MIL with SA~\cite{vaswani2017attention} & \xmark & 17.6K & 0.945 $\pm$ 0.035 & 0.853 $\pm$ 0.041 & 0.933 $\pm$ 0.063 & 0.816 $\pm$ 0.074 \\
      MIL with SA + axial PE~\cite{ramachandran2019stand} &  \smaller{ABS} & 17.6K & 0.941 $\pm$ 0.011 & 0.860 $\pm$ 0.063 & 0.963 $\pm$ 0.005 & 0.877 $\pm$ 0.015 \\
      MIL with SA + Fourier PE~\cite{yang2021learnable} &  \smaller{ABS} & 18.4K & 0.943 $\pm$ 0.025 & 0.874 $\pm$ 0.012 & 0.939 $\pm$ 0.034 & 0.863 $\pm$ 0.016 \\
      MIL with disc.\ rel.\ SA~\cite{wu2021rethinking} &  \smaller{REL} & 17.8K & 0.970 $\pm$ 0.005 & 0.975 $\pm$ 0.005 & 0.954 $\pm$ 0.031 & 0.923 $\pm$ 0.089 \\
      TransMIL~\cite{shao2021transmil} &  \smaller{ABS} & 2.18M & \textbf{0.991 $\pm$ 0.009} & 0.831 $\pm$ 0.041 & \textbf{0.993 $\pm$ 0.004} & 0.805 $\pm$ 0.046 \\
      DAS-MIL (ours) &  \smaller{REL} & 17.4K & 0.990 $\pm$ 0.003 & \textbf{0.992 $\pm$ 0.009} & 0.991 $\pm$ 0.002 & \textbf{0.970 $\pm$ 0.020} \\
      \bottomrule
    \end{tabular}
  }
\end{table}

\paragraph{Ablation study}
We report the results of the ablation study from \cref{sec:results:mnist} that investigated variants of the embeddings in \cref{tbl:ablation_embeddings}.
\begin{table}[h]
  \caption{%
    Results of varying the embeddings used in the DAS-MIL model. 
    The first column indicates which embeddings are used, \ie $\vb^K$ for key embeddings, $\vb^Q$ for query embeddings, and $\vb^V$ for value embeddings as in \cref{eq:das:compatibility,eq:das:output}.
    Unused embeddings are set to zero, \ie the model ``DAS-MIL ($\vb^K$)'' uses only key embeddings and sets $\vb^Q=\vb^V=\vzero$.
    In the fourth and eighth rows, we add the term $\vb_{ij}^Q(\vb_{ij}^K)^\top$ in the numerator of the compatibility function \cref{eq:das:compatibility:impl}.
    The penultimate row shows the performance of our method if all embeddings are kept fixed, \ie using non-trainable random vectors.
    The last row refers to the main DAS-MIL model described throughout the paper which uses all three trainable embeddings.
  }
  \label{tbl:ablation_embeddings}
  \centering
  \adjustbox{max width=\linewidth}{
    \begin{tabular}{l|r|rr|rr}
      \toprule
       &  & \multicolumn{2}{c|}{\smaller{MNIST-COLLAGE}} & \multicolumn{2}{c}{\smaller{MNIST-COLLAGE-INV}} \\
      Model  & \multicolumn{1}{c|}{Params} & \multicolumn{1}{c}{Train} & \multicolumn{1}{c|}{Test} & \multicolumn{1}{c}{Train} & \multicolumn{1}{c}{Test} \\
      \midrule
      DAS-MIL ($\vb^K$) & 17.3K & 0.929 $\pm$ 0.059 & 0.910 $\pm$ 0.029 & 0.870 $\pm$ 0.073 & 0.836 $\pm$ 0.077 \\
      DAS-MIL ($\vb^Q$) & 17.3K & 0.919 $\pm$ 0.014 & 0.914 $\pm$ 0.055 & 0.848 $\pm$ 0.108 & 0.836 $\pm$ 0.121 \\
      DAS-MIL ($\vb^V$) & 17.3K & 0.926 $\pm$ 0.017 & 0.912 $\pm$ 0.015 & 0.917 $\pm$ 0.030 & 0.890 $\pm$ 0.029 \\
      DAS-MIL ($\vb^K, \vb^Q$) ($+ \vb^Q{\vb^K}^\top$ in \cref{eq:das:compatibility:impl}) & 17.3K & 0.899 $\pm$ 0.051 & 0.906 $\pm$ 0.065 & 0.907 $\pm$ 0.043 & 0.850 $\pm$ 0.062 \\
      DAS-MIL ($\vb^K, \vb^Q$) & 17.3K & 0.918 $\pm$ 0.035 & 0.926 $\pm$ 0.029 & 0.927 $\pm$ 0.019 & 0.940 $\pm$ 0.029 \\
      DAS-MIL ($\vb^K, \vb^V$) & 17.3K & 0.940 $\pm$ 0.023 & 0.934 $\pm$ 0.019 & 0.948 $\pm$ 0.008 & \textbf{0.942 $\pm$ 0.028} \\
      DAS-MIL ($\vb^Q, \vb^V$) & 17.3K & 0.927 $\pm$ 0.015 & 0.916 $\pm$ 0.013 & \textbf{0.954 $\pm$ 0.012} & 0.912 $\pm$ 0.026 \\
      DAS-MIL ($\vb^K, \vb^Q, \vb^V$) ($+ \vb^Q{\vb^K}^\top$ in \cref{eq:das:compatibility:impl}) & 17.4K & 0.932 $\pm$ 0.029 & 0.942 $\pm$ 0.013 & 0.910 $\pm$ 0.048 & 0.898 $\pm$ 0.071 \\
      \hline
      DAS-MIL (non-trainable $\vb^K, \vb^Q, \vb^V$) & 17.3K & 0.937 $\pm$ 0.013 & 0.952 $\pm$ 0.023 & 0.932 $\pm$ 0.045 & 0.896 $\pm$ 0.065 \\
      \hline
      DAS-MIL ($\vb^K, \vb^Q, \vb^V$) & 17.4K & \textbf{0.955 $\pm$ 0.011} & \textbf{0.958 $\pm$ 0.013} & 0.953 $\pm$ 0.004 & 0.906 $\pm$ 0.034 \\
      \bottomrule
    \end{tabular}
  }
\end{table}

\section{Details of the \smaller{CAMELYON16} experiments}
\label{app:camelyon16}

We performed the feature extraction as detailed in \cref{sec:results:camelyon} once before training because it is computationally expensive and we chose the feature extractor to be non-trainable.
The entire feature extraction pipeline (\ie splitting the WSI into tiles, discarding tiles with too little tissue, performing stain normalisation, and extracting features) was performed using the open-source \texttt{end2end-WSI-preprocessing} tool available at \url{https://github.com/KatherLab/end2end-WSI-preprocessing}.

\end{document}